\documentclass[journal]{IEEEtran}
\usepackage{lineno,hyperref}
\usepackage{graphicx}
\usepackage{epstopdf}
\usepackage{bm}
\usepackage{multirow}
\usepackage{threeparttable}
\usepackage[subfigure]{tocloft}
\usepackage{subfigure}
\usepackage{amsmath}
\usepackage{amssymb}
\usepackage{algorithmic}
\usepackage{algorithm}
\usepackage{booktabs}   
\usepackage{xcolor}     

\ifCLASSINFOpdf
\else
\fi
\begin{document}

\title{ FAHCD-Net: Frequency-Adaptive Heatmap-Conditional Diffusion Networks for Robust Facial Landmark Detection }


\author{
Jun~Wan,
Jiwei~Hu,
Shengkai~Hu,
and~Qilu~Zhu
\IEEEcompsocitemizethanks{
    \IEEEcompsocthanksitem J. Wan, J. Hu, S. Hu, and Q. Zhu are with the School of Information Engineering, Zhongnan University of Economics and Law, Wuhan 430073, China (e-mail: junwan2014@whu.edu.cn; jiweihu@stu.zuel.edu.cn; shengkaihu@stu.zuel.edu.cn; qiluzhu@stu.zuel.edu.cn). Corresponding author: Jun Wan.
}
}

\maketitle
\begin{abstract}
	Facial Landmark Detection(FLD) is a crucial task in various applications and has achieved significant advancements in recent years.
	However, current FLD methods still struggle under challenging conditions, where facial structural variations, information loss, and noise interference severely compromise the integrity and accuracy of learned facial features.
	To address these issues, we propose Frequency-Adaptive Heatmap-Conditional Diffusion Network (FAHCD-Net), which integrates a Frequency-Adaptive Heatmap-Conditional Diffusion (FAHCD) model with a Smoothness Regularization (SR) loss in a cascaded framework.
	Specifically, the FAHCD model incorporates a Hierarchical Frequency Adaptation (HFA) module designed to suppress redundant high-frequency noise through multi-layer frequency decomposition and adaptive reconstruction, thereby preserving essential facial structures. Additionally, the SR loss is proposed to further mitigate the interference of high-frequency noise and enhance the smoothness of the generated landmark heatmaps. 
	By cascading the FAHCD model with the SR loss, FAHCD-Net effectively leverages both statistical and frequency-based distribution characteristics of the data to progressively generate more accurate landmark heatmaps from noisy inputs. Extensive experiments on popular benchmarks demonstrate the effectiveness and robustness of the proposed method, achieving state-of-the-art performance in FLD tasks under challenging scenarios.
	The source code is available at \url{https://github.com/HJWKryptonite/FAHCD-Net}.
\end{abstract}

\begin{IEEEkeywords}
Facial Landmark Detection, Diffusion Probabilistic Models, Heatmap regression
\end{IEEEkeywords}

\IEEEpeerreviewmaketitle

\section{Introduction}

\IEEEPARstart{F}{acial} Landmark Detection (FLD) is an important research direction in computer vision and image processing, 
involving the localization of specific landmarks in facial images (such as eye corners, nose tip, lips, and facial contour). 
This task is fundamental to various practical applications, including face recognition \cite{Chai2023RecognizabilityEE,hu3}, face synthesis \cite{Bao2018TowardsOI,hu6,wan10}, expression analysis \cite{Bettadapura2012FaceER,wan5}, and 3D face reconstruction \cite{Garrido2016ReconstructionOP}.

\begin{figure}[t]
	\centering 
	\includegraphics[width=0.98\linewidth]{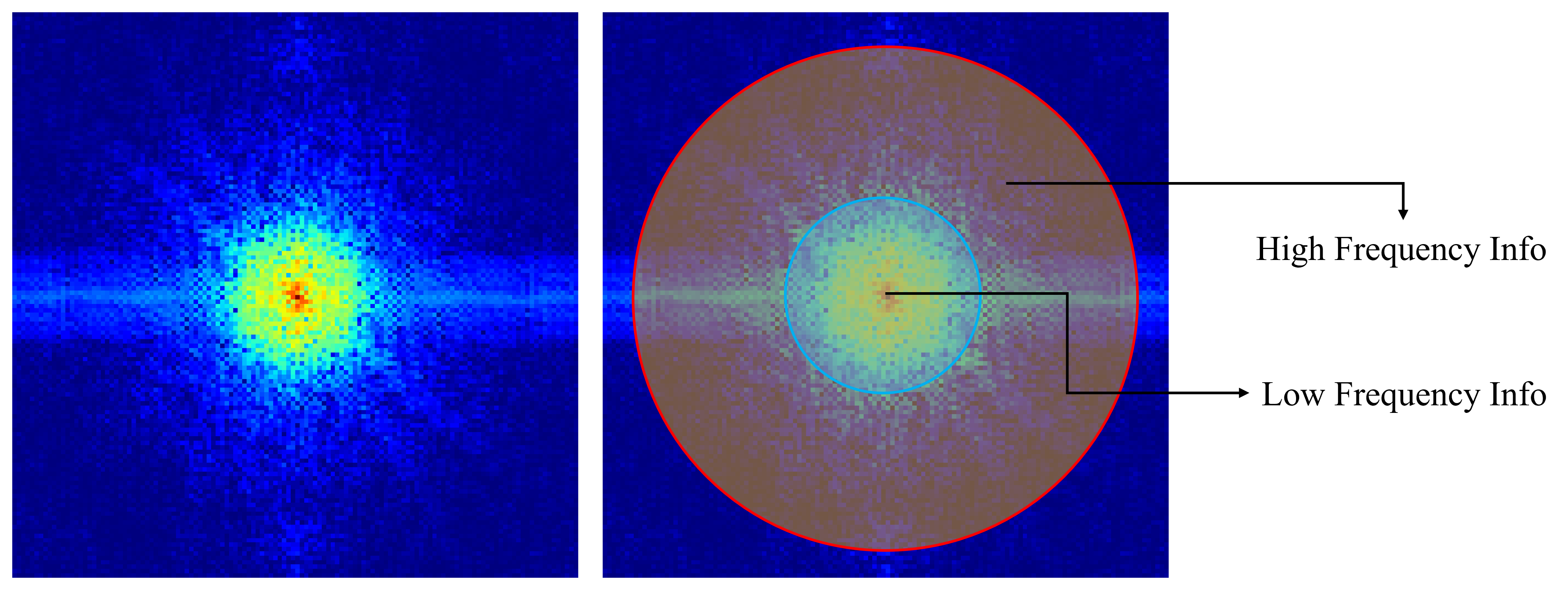}
	\vspace{-1em} 
	\caption{
		Composition of the Power Spectral Density (PSD) map. The area within the red circle indicates the main frequency information. 
		The yellow region shows the high-frequency components (i.e., faster-changing parts of the image, such as edges, textures, and details) of the image, 
		while the blue region represents the low-frequency components (i.e.,  slower-changing parts of the image, usually including large-scale backgrounds and smoother areas, such as cheeks and forehead).
	} 
	\label{fig1} 
	\vspace{-1em} 
\end{figure}

\begin{figure}[t]
	\centering 
	\includegraphics[width=0.98\linewidth]{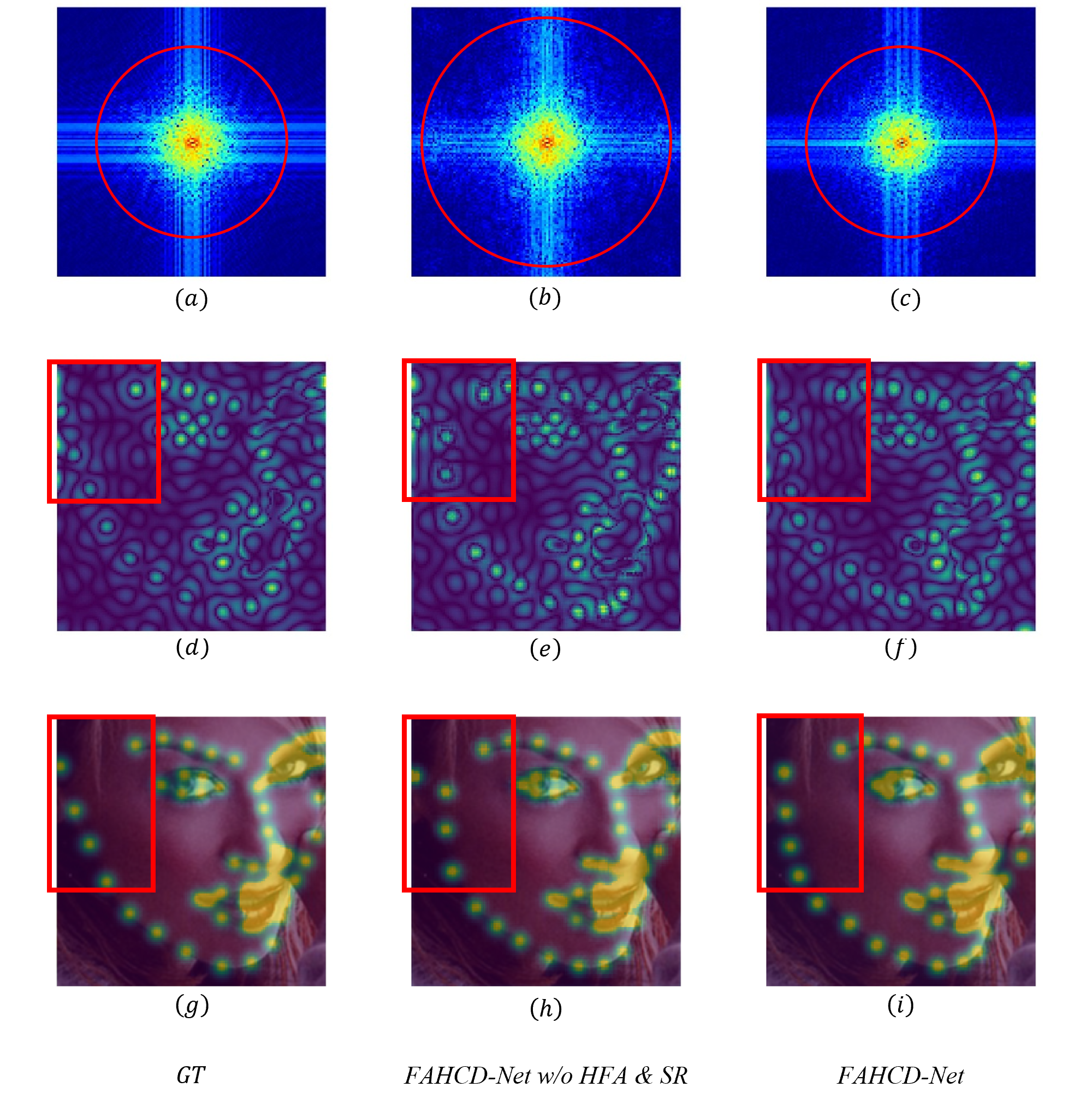}
	\vspace{-1em} 
	\caption{
		Visualization of Power Spectral Density (PSD) maps (the first row), High Frequency Component (HFC) maps (the second row) and landmark heatmaps (the last row). Red circles highlight the main frequency information, while red boxes mark the shifted landmarks and their corrections by our proposed FAHCD-Net. Observations reveal that heatmaps generated solely by the diffusion model (FAHCD model w/o HFA module) contain some high-frequency noise (i.e., red box in (e)), causing shifted landmarks (i.e., red box in (h)). In contrast, heatmaps generated by the FAHCD-Net effectively suppress excessive high-frequency noise (i.e., red box in (f)), resulting in a frequency distribution closer to the ground-truth (i.e., red box in (c)). Consequently, the landmark position is corrected. (as shown in (i)).
	} 
	\label{fig2} 
	\vspace{-1em} 
\end{figure}

With the development of convolutional neural network (CNN)\cite{wan2,zhu1,hu1,hu2,hu4,hu5}, FLD has made significant progress. Early methods for FLD were mostly based on coordinate regression approaches \cite{Sun2013DeepCN, Zhou2013ExtensiveFL, Trigeorgis2016MnemonicDM, Xiao2016RobustFL, Li2022RePFormerRP}, which aimed to learn the mapping from CNN features to landmark coordinates through fully connected layers. Recent works predominantly adopt the heatmap regression method \cite{Kowalski2017DeepAN, Yang2017StackedHN, Dapogny2019DeCaFADC, Dong2018SupervisionbyRegistrationAU, Wang2019AdaptiveWL}, predicting intermediate heatmaps for each landmark and decoding the landmark coordinates from them. The current state-of-the-art FLD has achieved very impressive results.

However, FLD remains challenging in complex scenes where facial expressions, poses, illumination, blur, and occlusions vary drastically. These challengings can be categorized into three aspects:
\textbf{(1) Facial structure variations}: Facial expressions and large pose variations can cause significant shifts in landmark positions and alterations in geometric relationships, increasing the complexity of landmark distributions. Current models often struggle to handle these variations effectively.
\textbf{(2) Information loss}: Occlusion and extreme poses (e.g., profile views) render some landmark information directly invisible, causing the loss of information.
\textbf{(3) Noise interference}: Blur and illuminations introduce substantial noise that can mess up the model's predictions.

Recently, Diffusion Probabilistic Models (DPMs) and score-based generative models \cite{Ho2020DenoisingDP, Rombach2021HighResolutionIS, Nichol2021GLIDETP} have achieved impressive results in image generation and processing, surpassing traditional GANs \cite{Goodfellow2014GenerativeAN}. Furthermore, the inherent characteristics of diffusion models enable them to perform effectively in addressing the three types of challenging scenarios mentioned above:
(1) Diffusion models learn the statistical distribution of data rather than relying solely on simple geometric features or predefined geometric constraints, which allows them to naturally adapt to variations in facial expressions and poses.
(2) The generative nature and multi-step recursive prediction mechanism of diffusion models enable them to capture the dynamic patterns of landmarks' diffusion even in cases of missing features.
(3) The denoising process can extract useful information from the noise and iteratively recover the data, as the model is originally trained using different levels of noise.

Building on these strengths, we introduce diffusion models and propose the Frequency-Adaptive Heatmap-Conditional Diffusion Network (FAHCD-Net) for more robust FLD. As illustrated in Fig.~\ref{fig1}, facial images exhibit distinct power spectral density (PSD) distributions, where high-frequency components capture edges and fine textures while low-frequency components correspond to smooth regions such as cheeks and forehead. The FAHCD-Net progressively generates accurate facial landmark heatmaps from noise while constraining the frequency distribution and capturing complex structural details. Moreover, to reduce the impact of redundant high-frequency noise on the quality of generated landmark heatmaps, we also design a Frequency-Adaptive Heatmap-Conditional Diffusion (FAHCD) model and a Smoothness regularization (SR) loss. The former introduces a Hierarchical Frequency Adaptation (HFA) module to perform frequency decomposition and reconstruction, adaptively adjusting the proportion of different frequency components, and the latter imposes smoothness and continuity constraints on the generated heatmap by penalizing dramatically changed parts of landmark heatmaps. As shown in Fig.~\ref{fig2}, the diffusion model without our proposed modules generates heatmaps with excessive high-frequency noise, leading to shifted landmark predictions. In contrast, FAHCD-Net effectively suppresses such noise and corrects the landmark positions. Therefore, by cascading the FAHCD model and SR loss, the proposed FAHCD-Net outperforms state-of-the-art FLD methods and its main contributions are as follows:

\begin{enumerate}
    \item Building on the importance of frequency information in modeling facial structures and leveraging the advantages of diffusion models in learning statistical distribution features and noise robustness, we propose FAHCD model to cope with FLD in challenging scenarios.
    
    \item A well-designed SR loss is proposed to suppress redundant high-frequency noise and constrain the generated landmark heatmap to ensure its smoothness and continuity, thereby achieving accurate FLD, especially for occluded, illuminated and blurred faces.
    
    \item A novel framework called FAHCD-Net is developed to seamlessly integrate the FAHCD model and SR loss in a cascaded manner to handle FLD in challenging scenarios. Experimental results show that FAHCD-Net outperforms the state-of-the-art methods on multiple challenging datasets (such as 300W, COFW, WFLW, and AFLW).
    
\end{enumerate}

\section{Related Work}

In this section, we will introduce the related work on facial landmark detection and diffusion models.

\subsection{Facial Landmark Detection}

In the early stages, FLD is primarily based on statistical model methods, such as Active Appearance Models (AAM) \cite{927467}, Active Shape Models (ASM) \cite{COOTES199538}, and Constrained Local Models (CLM) \cite{Cristinacce2006FeatureDA}. Recently, with the development of CNNs\cite{wan4,wan9,wan8,wan6,wan3}, deep learning methods have achieved significant success. Deep learning-based facial landmark detection mainly follows two mainstream approaches: coordinate regression methods \cite{Li2022RePFormerRP, Yin2024SCEMAESC,  Liang2024GeneralizableFL,wan7,WAN13} and heatmap regression methods \cite{Tourani2024PoseGuidedSW, Jin2020PixelinPixelNT,wan4,WAN14}.

\textbf{Coordinate Regression-Based Methods.} These methods primarily use fully connected layers to learn the mapping between facial features and landmark coordinates. For example, Sun et al. \cite{Sun2013DeepCN} are the first to apply CNNs to facial landmark detection, proposing a deep convolutional neural network based on cascaded regression, which achieves more accurate landmark detection results. Later, to further enhance performance, MDM \cite{Trigeorgis2016MnemonicDM} and RAR \cite{Xiao2016RobustFL} employ cascaded recurrent refinements to sequentially fine-tune landmark estimation. Li et al. \cite{Li2020StructuredLD} propose a new topology-adaptive deep graph to strengthen the results. SLPT \cite{Xia2022SparseLP} and RePFormer \cite{Li2022RePFormerRP} employ attention mechanisms to learn an adaptive inherent relationship. SCE-MAE \cite{Yin2024SCEMAESC} addresses the limitations of self-supervised learning in facial landmark detection tasks by combining the pretraining advantages of MAE with a selective optimization strategy for corresponding relationships. Liang et al. \cite{Liang2024GeneralizableFL} combine conditional facial deformation with landmark detection to learn a highly generalized landmark detector through an alternating optimization approach, significantly improving detection performance on stylized faces and unseen data. Although coordinate regression methods are simple and efficient, they often struggle to capture complex spatial relationships and local details, particularly when handling large pose variations and partial occlusions.

\textbf{Heatmap Regression-Based Methods.} These methods output an intermediate heatmap for each landmark and use the Argmax function to consider the point with the highest intensity as the best output. For example, Kowalski et al. \cite{Kowalski2017DeepAN} introduce a novel cascaded deep neural network called DAN (Deep Alignment Network), which stands out from earlier cascaded networks by utilizing the whole image as input at every stage instead of focusing on specific image regions. SBR \cite{Dong2018SupervisionbyRegistrationAU} utilizes the registration of synthetic images to provide supervision signals for training. By analyzing the main shortcomings of different loss functions, AWingLoss \cite{Wang2019AdaptiveWL} is proposed to address the imbalance between foreground and background pixels. HRNet \cite{Wang2019DeepHR} connects and exchanges information by merging multi-scale image features from multiple branches, thereby generating more effective heatmaps. More recently, PIPNet \cite{Jin2020PixelinPixelNT} proposed performing heatmap and offset predictions concurrently on low-resolution feature maps, significantly improving inference efficiency while maintaining competitive accuracy. LDEQ \cite{Micaelli2023RecurrenceWR} utilizes cascaded computation based on DEQ and achieve state-of-the-art performance. STARLoss \cite{Zhou2023STARLR} is proposed to cope with the semantic ambiguity problem in order to improve the FLD accuracy. PoseGuidedSW \cite{Tourani2024PoseGuidedSW} innovatively integrates diffusion model features, a self-training mechanism, a pose-guided proxy task, and a two-stage clustering approach, providing an efficient and robust solution for unsupervised landmark detection tasks.

However, heatmap regression methods rely on the quality of the heatmaps, where low quality may lead to blurred details and inaccurate localization. In contrast, diffusion models generate high-quality heatmaps through a step-by-step denoising process, enabling them to recover landmark details more precisely.

\begin{figure*}[t]
	\centering 
	\includegraphics[width=18cm,height=10cm]{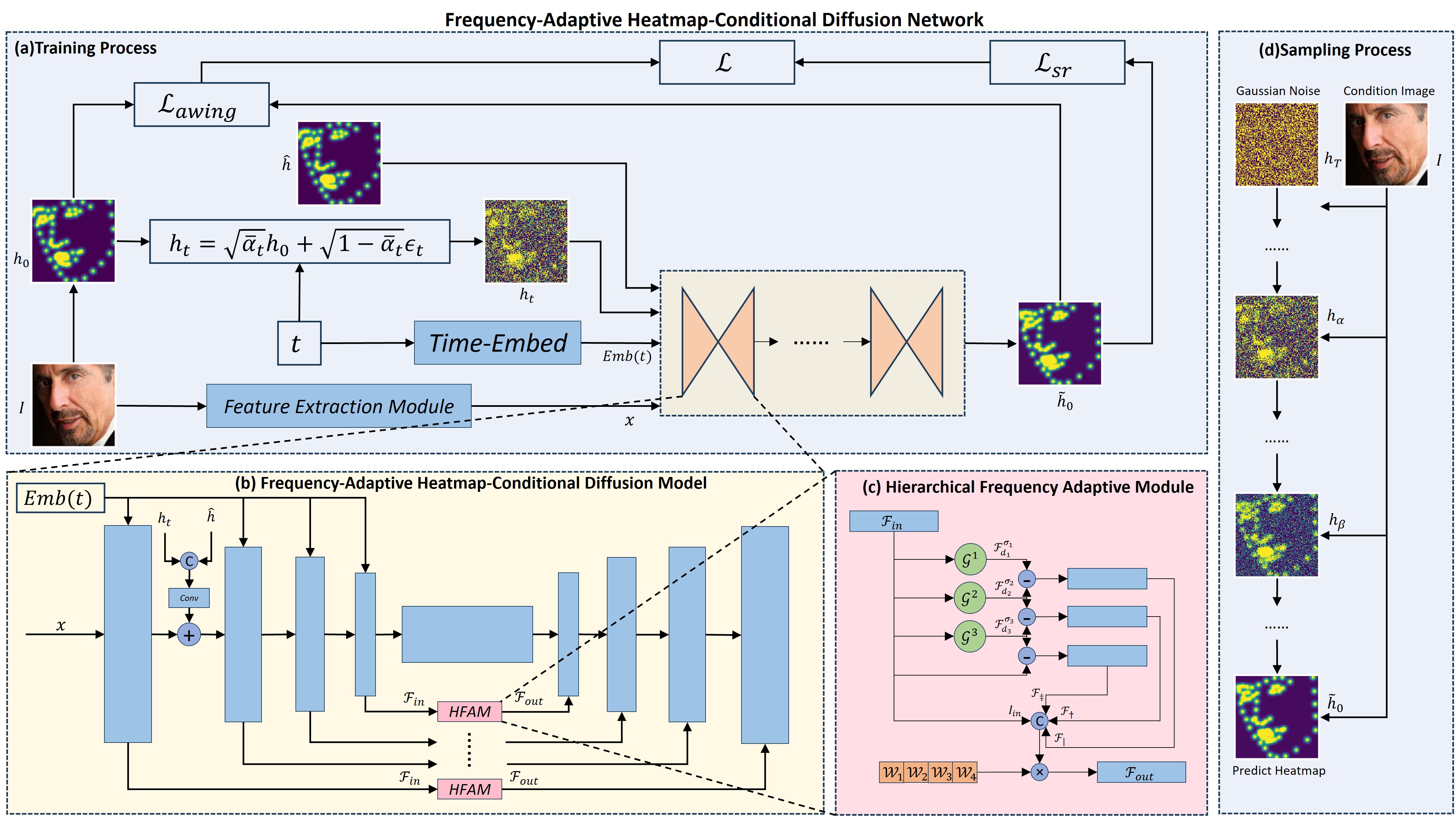}
	\caption{
		The overview of our proposed FAHCD-Net. (a) Training Process, (b) FAHCD model, (c) HFA module, and (d) Sampling Process. It consists of FAHCD model and SR loss. The FAHCD model integrates HFA modules to perform frequency decomposition and reconstruction, adaptively adjusting the contribution of different frequency components.	SR loss aims to apply smoothness and continuity constraints to the generated heatmaps. Then, by integrating the FAHCD model and SR loss in a cascading manner, the proposed FAHCD-Net can achieve outstanding performance.
	} 
	\label{fig3} 
	\vspace{-1em} 
\end{figure*}

\subsection{Diffusion Models}

Diffusion Probabilistic Models (DPMs) are a type of generative model based on a Markov chain. These models can transform noise sampled from a simple distribution (e.g., Gaussian distribution) into target data sampled under a complex distribution. Ho et al. systematically explain the denoising diffusion probabilistic model (DDPM) \cite{Ho2020DenoisingDP}, demonstrating the outstanding performance in image generation, surpassing GANs \cite{Goodfellow2014GenerativeAN} in many tasks. Nichol and Dhariwal address the low sampling efficiency issue of diffusion models by adjusting the diffusion process, further improving the quality of image generation. Latent Diffusion Models (LDM) \cite{Rombach2021HighResolutionIS} significantly reduces computational resource requirements by running the diffusion process in latent space, making them especially effective in high-resolution image generation tasks. DDNM \cite{Wang2022ZeroShotIR} formulates a sophisticated identity equation that seamlessly integrates conditions into the reverse process of the diffusion model without requiring additional training, demonstrating excellent performance in linear image restoration tasks. Conditional Diffusion Models (CDM)\cite{Niu2023CDPMSRCD} enhance control over the generative process by introducing conditional variables. The EDM\cite{Karras2022ElucidatingTD} framework decomposes the design of complex diffusion models into key components: the diffusion process, model architecture, optimization objectives, and sampling procedures. This modular approach provides researchers with a clearer understanding of these models and facilitates their improvement. DPM-solver\cite{Lu2022DPMSolverAF} further accelerates sampling by employing methods like calculating the exact ODE solutions and designing higher-order solvers. Currently, extensive research and applications are being explored in areas such as controllable image generation (ControlNet\cite{Zhang2023AddingCC}), image editing (DreamBooth\cite{Ruiz2022DreamBoothFT}), image inpainting (SmartBrush\cite{Xie2022SmartBrushTA}), and style transfer (StyleDrop\cite{Sohn2023StyleDropTG}). Given the powerful image generation capabilities of the diffusion model in the tasks mentioned above, we introduce it to enhance landmark heatmap generation, aiming to tackle the challenges posed by FLD.

\section{Methodology}
In this section, we introduce our proposed FAHCD-Net, as depicted in Fig. \ref{fig3}. We start by outlining the preliminaries of DDPM in Section III.A, followed by a comprehensive overview of FAHCD-Net in Section III.B. Detailed explanations of the FAHCD model and the SR loss are provided in Sections III.C and III.D, respectively.

\subsection{Preliminary}

In the diffusion model, given an initial data (e.g., image) distribution $x_0 \sim q(x)$, Gaussian noise is gradually added to the distribution. Given a predefined variance schedule $\{\beta_t\}_{t=1}^T$, noise is added at each step to the previous data $x_{t-1}$ as follows:

\begin{small}
\begin{equation}
	q(x_t|x_{t-1}) = \mathcal{N}(x_t; \sqrt{1-\beta_t}x_{t-1}, \beta_t I)
\end{equation}
\end{small}

Then, with the reparameterization trick, the noise distribution at any time-step $t$ can be computed as $q(x_t|x_0)$:

\begin{small}
\begin{equation}
	q(x_t|x_0) = \mathcal{N}(\sqrt{\overline\alpha_t}x_0, (1-\overline\alpha_t)I), \quad \overline\alpha_t = \prod_{i=1}^t(1-\beta_i)
\end{equation}
\vspace{-1em}
\end{small}

Based on this distribution, $x_t$ at a specific time-step $t$ can be computed directly from $x_0$ as follows:

\begin{small}
\begin{equation}\label{forward}
	x_t=\sqrt{\bar{\alpha}_t}x_0+\sqrt{1-\bar{\alpha}_t}\epsilon_t	
\end{equation}
\end{small}where \(\epsilon_t\) follows a standard Gaussian distribution \(\mathcal{N}(0, I)\), and \(\epsilon_t \in \mathbb{R}^{3 \times h \times w}\).

The reverse process $q(x_{t-1}|x_t)$ in the diffusion model is designed to progressively reconstruct data from noise, leveraging a deep neural network to model the state transition from $x_t$ to $x_{t-1}$. The distribution $p_\theta(x_{t-1}|x_t)$, representing this transition, can be expressed as follows:

\begin{small}
\begin{equation}
	p_\theta(x_{t-1}|x_t)=\mathcal{N}(x_{t-1};\mu_\theta(x_t,t),\Sigma_\theta(x_t,t))
\end{equation}
\end{small}where $\Sigma_\theta(x_t,t)$ is usually a predefined constant related to the variance schedule \cite{Ho2020DenoisingDP}, and $\mu_\theta(x_t, t)$ is typically parameterized by a denoising network $\epsilon_\theta(x_t, t)$ according to:

\begin{small}
\begin{equation}
	\mu_\theta(x_t,t) = \frac{1}{\sqrt{\alpha_t}} \left( x_t - \frac{1-\alpha_t}{\sqrt{1-\overline\alpha_t}} \epsilon_\theta(x_t,t) \right)	
\end{equation}
\end{small}

Generally, the diffusion model is supervised by the $\mathcal{L}_\text{simple}$ loss function with respect to $\theta$:

\begin{small}
\begin{equation}
	\mathcal{L}_\text{simple} = \sum_{t=1}^T \left[ \| \epsilon_\theta(x_t,t) - \epsilon_t \|_2^2 \right]
\end{equation}
\end{small}

\subsection{FAHCD-Net Overview}

The denoising process of FAHCD-Net takes facial image features $x$ as inputs and uses conditional landmark heatmap $\hat{h}$ as conditional information, which can be defined as follows:

\begin{small}
\begin{equation}
	p_\theta(h_{0:T}|x,\hat{h}) = p(h_T) \prod_{t=1}^T p_\theta(h_{t-1}|h_t,x,\hat{h})	
\end{equation}
\end{small}where $h_t$ denotes the noisy landmark heatmap correspond to time-step $t$, and $h_0$ represents the ground-truth landmark heatmap.

The conditional distribution $p_\theta(h_{t-1}|h_t,x,\hat{h})$ is given by:

\begin{small}
\begin{equation}
	p_\theta(h_{t-1}|h_t,x,\hat{h}) =
	\begin{cases}
		\mathcal{N}(\mu_\theta(h_1,1,x,\hat{h}), 0), & \text{if } t=1 \\
		q(h_{t-1}|h_t,\mu_\theta(h_t,t,x,\hat{h})), & \text{otherwise}
	\end{cases}
\end{equation}
\end{small}

Following the DDIM \cite{Song2020DenoisingDI}, we employ a deterministic generation process:

\begin{small}
\begin{equation}
	\mu_\theta(h_t,t,x,\hat{h}) = \frac{1}{\sqrt{\alpha_t}} \left( h_t - \sqrt{1-\alpha_t} \epsilon_\theta(h_t,t,x,\hat{h}) \right)
\end{equation}
\end{small}where $\epsilon_\theta$ denotes the proposed FAHCD model. Note that FAHCD-Net does not predict the noise directly but instead predicts the intermediate heatmaps. And the heatmap prediction formula can be derived as:

\begin{small}
\begin{equation}
	\tilde{h}_0=\frac{h_t-\sqrt{1-\bar{\alpha_t}}\epsilon_t}{\sqrt{\bar{\alpha_t}}}
\end{equation}
\end{small}where $\tilde{h}_0$ denotes the generated landmark heatmap.

In the experiments, we found that using MSE or AWingLoss \cite{Wang2019AdaptiveWL,wan11,WAN12} to optimize the FAHCD-Net led to local mutations and redundant high-frequency noise in the heatmap. To tackle this issue, we further propose SR loss to guarantee the smoothness and continuity of the generated heatmaps. Accordingly, the complete loss function is defined as follows:

\begin{small}
\begin{equation}
	\mathcal{L}=\mathcal{L}_{awing}+\lambda\mathcal{L}_{sr}	
\end{equation}
\end{small}where $\lambda$ is used to balance these two losses.

Inspired by \cite{Ho2021CascadedDM}, we cascade multiple FAHCD-Nets across three stages to gradually enhance the effectiveness of generated landmark heatmap, which can be formulated as follows:

\begin{small}
	\begin{equation}
		\tilde{h}_0^s = \text{FAHCD-Net}^s(h_t, t, x, \hat{h}), s \in \{0, 1, 2\}
	\end{equation}
\end{small}where $s$ denotes the index of the current stage, with $s \in \{0, 1, 2\}$. When $s=0$, $\hat{h}^s$ represents the conditional landmark heatmap generated by current FLD models (e.g., SHN \cite{Yang2017StackedHN}). For $s>0$, $\hat{h}^s$ corresponds to the generated landmark heatmap from the previous stage (i.e., $\hat{h}^s=\tilde{h}_0^{s-1}$). Therefore, by cascading the FAHCD-Net, richer conditional facial priors can be leveraged to generate more accurate landmark heatmaps, thus achieving precise landmark detection.

Next, we will introduce the proposed FAHCD model and SR loss in detail.

\begin{figure}[t]
	\centering 
	\includegraphics[width=\columnwidth]{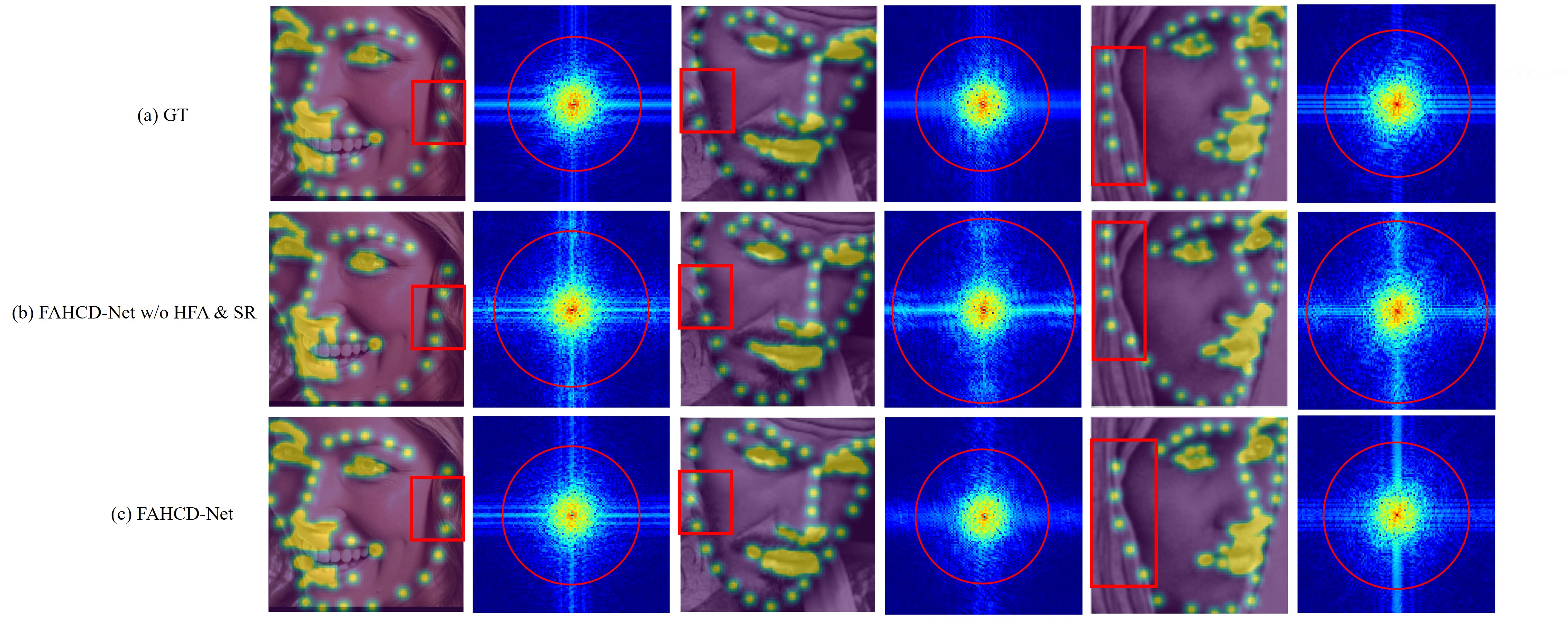}
    \vspace{-1em} 
	\caption{
		Comparison of heatmaps and their corresponding Power Spectral Density (PSD) maps. The first, second and third rows show landmark heatmap and PSD maps generated by ground-truth, FAHCD-Net w/o HFA module and SR loss, and FAHCD-Net, respectively. From the results, it can be observed that the heatmap generated by using FAHCD-Net w/o HFA module and SR loss produces a significant amount of additional high-frequency noise compared to the ground-truth. These noises appear as strong high-frequency components in the PSD map, affecting the precise localization of landmarks. In contrast, FAHCD-Net effectively suppresses unnecessary high-frequency noise by combining HFA module and SR loss, thereby improving the quality of the heatmap and the accuracy of landmark detection.
	} 
	\label{fig4} 
	\vspace{-1em} 
\end{figure}

\subsection{FAHCD model}

Currently, most DPMs \cite{Ho2020DenoisingDP, Rombach2021HighResolutionIS, Niu2023CDPMSRCD} are based on the U-net framework. When applied to FLD tasks, it is necessary to generate landmark heatmaps step-by-step by learning the pixel-level difference from Gaussian noise to the target heatmap. However, during this generation process, the frequency information of images undergoes significant changes(as shown in Fig.\ref{fig4}), which profoundly degrades the quality of the generated heatmaps for landmark detection. 

In image processing, different frequency components typically reflect specific characteristics of an image: high-frequency signals represent rapidly changing features such as edges and textures, while low-frequency signals correspond to more gradually varying features, such as smooth regions. For facial images, high-frequency signals are primarily concentrated in key facial feature areas (e.g., eyes, nose, mouth) and facial contours, whereas low-frequency signals are more likely to appear in smooth facial regions and the background.

Building on this, we further analyzed the frequency components of the ground-truth heatmap $h_0$ and the generated heatmap $\tilde{h}_0$, revealing significant differences between the two (as shown in Fig. \ref{fig4}). Specifically, low-frequency information remains stable, but high-frequency information increases in the generated heatmap compared to the ground-truth heatmap. This suggests that while diffusion models preserve the overall structure of the image, they also introduce additional high-frequency information during the generation process, which may cause the following problems:
(1) Peak blurring: The clarity of the Gaussian distribution is disrupted, reducing localization accuracy.
(2) Noise overfitting: The model may overfit the high-frequency noise in the training data, leading to diminished generalization capability.
(3) Noise in smooth areas: Even in the smooth regions of the heatmap, the presence of noise can interfere with the accuracy of landmark localization. Irrelevant noise in these areas may mislead the model into producing incorrect activations in non-target regions.
Therefore, an ideal landmark heatmap should exhibit a clear Gaussian peak at key positions and remain smooth in other regions. To achieve this, we propose the FAHCD model. FAHCD model is actually an hourglass unit, which is a multi-scale feature framework. So we introduce Hierarchical Frequency Adaptive (HFA) module at each scale of the FAHCD model.

In HFA module, the multi-scale features undergo frequency decomposition and reconstruction to better align with the ground-truth heatmap's frequency distribution. For frequency decomposition, the HFA module uses multi-scale Gaussian blur (i.e., Gaussian kernels with different sizes and standard deviations) as low-pass filters to achieve hierarchical frequency processing and feature filtering. For frequency reconstruction, we introduce a trainable parameter $W$ to combine decomposed frequency components. The design of the HFA module is illustrated in Fig. \ref{fig3}(c).

Assume that $\mathcal{K}^{\sigma}_{d\times d}$ represents a 2D Gaussian kernel with a kernel size of $d$ and a standard deviation of $\sigma$. Hence, a Gaussian blur process can be formulated as:

\begin{small}
	\begin{equation}
		\mathcal{F}_{d}^{\sigma}=\mathcal{K}^{\sigma}_{d\times d}*\mathcal{F}_{in}	
	\end{equation}
\end{small}where $*$ denotes the convolution operation, and $d\in\{3,5,7,\ldots\}$. $\mathcal{F}_{in}$ signifies the input features and $\mathcal{F}_{d}^{\sigma}$ represents the output after Gaussian blurring.

Three different Gaussian kernels with varying sizes and standard deviations are applied to filter varying frequency components (i.e., high-frequency component $\mathcal{F}_\lvert$, mid-frequency component $\mathcal{F}_\dag$ and low-frequency component $\mathcal{F}_\ddag$). They will be concatenated with $\mathcal{F}_{in}$ and then fused to obtain the filtered feature $\mathcal{F}_{out}$:

\begin{small}
	\begin{equation}
		[\mathcal{F}_{in},\mathcal{F}_\lvert,\mathcal{F}_\dag,\mathcal{F}_\ddag]=[\mathcal{F}_{in},\mathcal{F}_{in}-\mathcal{F}_3^{\sigma_1},\mathcal{F}_3^{\sigma_1}-\mathcal{F}_5^{\sigma_2},\mathcal{F}_5^{\sigma_2}-\mathcal{F}_7^{\sigma_3}]
	\end{equation}
\end{small}

\begin{small}
	\begin{equation}
		\mathcal{F}_{out}=W[\mathcal{F}_{in},\mathcal{F}_\lvert,\mathcal{F}_\dag,\mathcal{F}_\ddag]^T
	\end{equation}
\end{small}where $W=[w_1,w_2,w_3,w_4]$ is a trainable parameter that can be optimized during the training process.

Since the FAHCD is based on the U-net framework, which is the encoder-decoder structure, $\mathcal{F}_{out}$ will be concatenated with the corresponding original encoder output into the Decoder. Therefore, the reconstruction frequency information is introduced to eliminate redundant noise, thus modeling more effective facial structure and obtaining more accurate landmark detection. The complete training process is presented in Alg. \ref{alg:1}.

\begin{algorithm}[t]
    \caption{FAHCD-Net Training Process}
    \label{alg:1}
    \renewcommand{\algorithmicrequire}{\textbf{Require:}}
    \begin{algorithmic}[1]
        \REQUIRE Max Diffusion Step $T$, Image Set $q(x)$, ground-truth heatmap $h$, conditional heatmap $\hat{h}$ and hyper-parameter $s=0.008$
		\FOR{each input sample $x_0\in q(x_0)$}
        \STATE Sample $\epsilon\sim\mathcal{N}(0,I$)
		\STATE Sample $t\sim\mathcal{U}(\{1,\cdots,T\})$
     	\STATE $\beta_t=1-\frac{cos((t/T+s)/(1+s)\cdot\pi/2)}{cos(s\cdot\pi/2)}$
     	\STATE $\alpha_t=1-\beta_t$
     	\STATE $\bar{\alpha}_t=\prod_{s=0}^t{\alpha}_s$
     	\STATE Calculate $h_t=\sqrt{\bar{\alpha}_t}h+\sqrt{1-\bar{\alpha}_t}\epsilon$
     	\STATE Predict noise: $\epsilon_\theta(h_t,t,x,\hat{h})$
	 	\STATE Predict output heatmap $\tilde{h}_0$: \\
			$\mu_\theta(h_t,t,x,\hat{h}) = \frac{1}{\sqrt{\alpha_t}} \left( h_t - \sqrt{1-\alpha_t} \epsilon_\theta(h_t,t,x,\hat{h}) \right)$ \\
			$\tilde{h}_0=\frac{h_t-\sqrt{1-\bar{\alpha_t}}\epsilon_t}{\sqrt{\bar{\alpha_t}}}$
	 	\STATE Calculate the heatmap loss term $\mathcal{L}_{awing}$
	 	\STATE Calculate the Smoothness Regularization term $\mathcal{L}_{sr}$
	 	\STATE Update with $\mathcal{L}=\mathcal{L}_{awing}+\lambda\mathcal{L}_{sr}$
        \STATE \textbf{Until} convergence
		\ENDFOR
    \end{algorithmic}
\end{algorithm}

\subsection{Smoothness Regularization Loss}

In addition to proposing the FAHCD model to incorporate frequency information for modeling facial structure and generating landmark heatmaps, we also introduce the SR loss to ensure the smoothness and continuity of the generated heatmaps.

SR loss can avoid excessive high-frequency noise interference with either Total Variation (TV) regularization or Heatmap Gradient (HG) regularization. The former is typically used to remove noise while preserving the overall structure of the image. By penalizing large changes in the output heatmap, it suppresses the generation of high-frequency noise while retaining the main edge information. The form of TV regularization is as follows:

\begin{small}
	\begin{equation}
		\mathcal{L}_{TV}(\tilde{h}_0)=\sum_{i,j}\left((\tilde{h}_0^{i+1,j}-\tilde{h}_0^{i,j})^2+(\tilde{h}_0^{i,j+1}-\tilde{h}_0^{i,j})^2\right)
	\end{equation}
\end{small}where $i,j$ represent the pixel indexes.

HG regularization directly adds constraints on the heatmap gradients, penalizing excessive gradient changes, which can be formulated as:

\begin{small}
	\begin{equation}
		\mathcal{L}_{HG}(\tilde{h}_0)=\sum_{i,j}\left((\frac{\partial \tilde{h}_0}{\partial x})^2+(\frac{\partial \tilde{h}_0}{\partial y})^2\right)
	\end{equation}
\end{small}

By computing the squared derivatives of the heatmap $\tilde{h}_0$ in both the horizontal and vertical directions, HG regularization suppresses large variations, ensuring the smoothness of the generated landmark heatmap. Note that our proposed SR loss can be either $\mathcal{L}_{TV}$ or $\mathcal{L}_{HG}$, and we have reported corresponding experimental results in Section IV.

\section{Experiments}
\begin{figure}[t]
	\centering
	\includegraphics[width=\columnwidth]{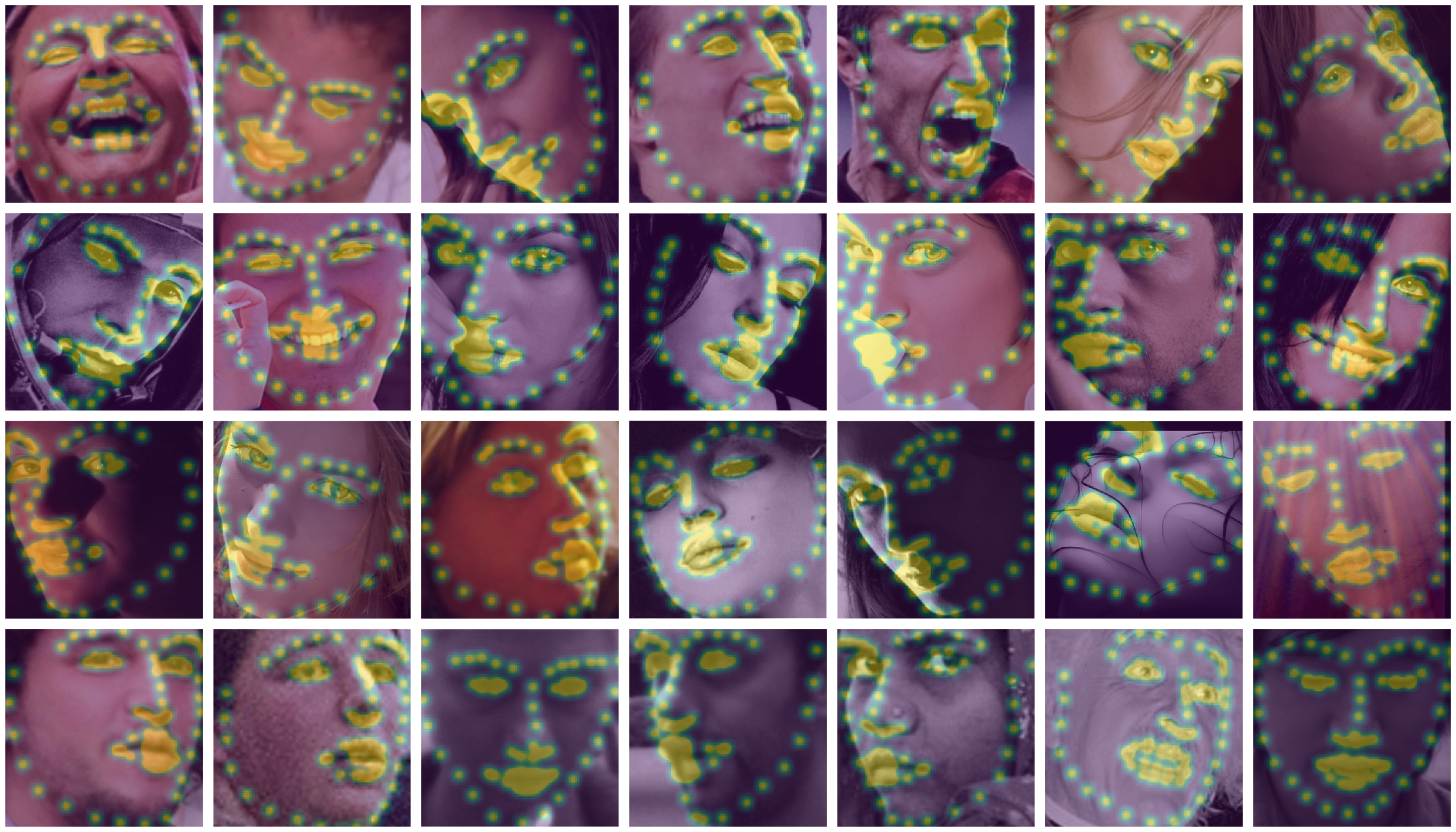}
	\centering
    \vspace{-2em}
	\caption{
		The experimental results of our proposed FAHCD-Net are shown in the figure. These examples demonstrate the performance of the model in various scenarios, illustrating its effectiveness in handling complex tasks. 
		The first row is images in facial expressions and large poses. The second row is in heavy occlusion. The third row is in illumination, and the last row is blur images.
	}
	\label{fig5}
	\vspace{-1em}
\end{figure}

In this section, we conduct experiments on our FAHCD-Net across multiple datasets to validate its effectiveness. Fig.~\ref{fig5} presents qualitative results across four challenging scenarios: large facial expressions and poses, heavy occlusion, illumination variation, and blur. The details are given as follows.

\subsection{Datasets and Experimental Settings}

\textbf{Datasets.} We evaluate the proposed methods on public datasets including 300W \cite{SAGONAS20163}, COFW \cite{BurgosArtizzu2013RobustFL}, WFLW \cite{Wu2018LookAB} and AFLW \cite{Zhu2016UnconstrainedFA}.

\textbf{300W}(68 landmarks): is a combination of LFPW, HELEN, AFW, and IBUG, with each face annotated with 68 landmarks. Typically, the training set consists of all images from AFW, as well as the training images from LFPW and HELEN, resulting in a total of 3,148 images. The test set includes all images from IBUG and the test images from LFPW and HELEN, amounting to 689 images. The LFPW and HELEN test images are designated as the Common Subset, whereas the IBUG images constitute the Challenging Subset.

\textbf{COFW}(29 landmarks): primarily focuses on images with occlusions. Typically, the training set consists of 845 faces from the LFPW training set and an additional 500 heavily occluded faces. The test set includes 507 images with significant occlusion. Evaluations are conducted on 29 landmarks.

\textbf{WFLW}(98 landmarks): consists of 10,000 facial images, with 7,500 allocated for training and 2,500 for testing, each annotated with 98 landmarks. Beyond landmark annotations, it provides extensive attribute information, including occlusion, pose, makeup, illumination, blur, and expression.

\textbf{AFLW} (19 landmarks): This dataset consists of 25,993 high-quality facial images, each annotated with 19 landmarks. It includes a diverse range of scenarios, featuring both indoor and outdoor scenes. Moreover, it covers challenging cases such as side profiles and various rotation angles, making it ideal for evaluating robust facial landmark detection methods.

\textbf{Evaluation metric.} Normalized Mean Error (NME) is a commonly used metric for evaluating landmark detection performance, which is defined as:

\begin{small}
	\begin{equation}
		\text{NME}(P,\hat{P})=\frac{1}{N_{P}}\sum_{i=1}^{N_P}\frac{\|p_i-\hat{p}_i\|_2}{d}
	\end{equation}
\end{small}where $P$ and $\hat{P}$ represent the ground-truth and predicted coordinates, respectively, while $p_i$ and $\hat{p}_i$ correspond to the coordinates of the $i$-th landmark in the ground-truth and prediction. $N_p$ represents the total number of landmarks, and $d$ serves as the reference distance to normalize the absolute errors. Commonly, the inter-ocular distance (distance between the outer corners of the eyes) or the inter-pupil distance (distance between the centers of the pupils) is used as the normalization factor.

Failure Rate (FR) is another important metric for evaluating landmark detection performance, which is defined as:

\begin{small}
	\begin{equation}
		\text{FR}(P,\hat{P}) = \frac{1}{N_P} \sum_{i=1}^{N_P} \mathbb{1} \left(\frac{\|p_i - \hat{p}_i\|_2}{d} > \tau \right)
	\end{equation}
\end{small}where the threshold $\tau$ defines the maximum normalized error allowed for a successful prediction, and $\mathbb{1}(\cdot)$ is an indicator function that returns 1 if the condition is true and 0 otherwise. FR calculates the proportion of landmarks whose normalized error exceeds the threshold, providing insight into the robustness of the detection method in handling difficult cases.

\textbf{Implementation Details.} 
The proposed FAHCD-Net is trained on an NVIDIA GeForce RTX 4090 GPU. For data preprocessing, all input images are uniformly cropped and resized to a spatial resolution of $256 \times 256$ pixels for both training and evaluation.

To enhance the robustness of the model, various image augmentation strategies are applied to the training set samples: (1) random clockwise or counterclockwise rotation between 0° and 18°; (2) random cropping of 0-5\%; (3) random grayscale conversion for 20\% of the samples; (4) random blurring for 30\% of the samples; (5) random occlusion for 40\% of the samples; (6) random horizontal flipping for 50\% of the samples.

The training process utilizes the Adam optimizer, starting with an initial learning rate of $1\times10^{-3}$. This rate is decayed by a factor of 0.9 at the 80th, 150th, and 200th epochs. The model is trained over 300 epochs with a batch size of 12.  

In terms of hyperparameter settings, $\omega$ for $\mathcal{L}_{heatmap}$ is set to 14, $\theta$ to 0.5, $\epsilon$ to 1, and $\alpha$ to 2.1. The weight for the smoothness regularization term is set to $\lambda=3\times10^{-7}$. When processing the dataset, a Gaussian kernel ($\sigma=3$) is empirically used to generate the landmark heatmaps corresponding to the images.

\subsection{Evaluations under Normal Circumstances}

Under normal circumstances, we primarily conducts comparative experiments on the 300W Common Subset, 300W Fullset and AFLW-frontal subset. These subsets consist mostly of facial images captured in favorable conditions, with clear images and minimal pose variations, illuminations, occlusion, or noise interference.

Tab. \ref{tab:300w_sota} presents the comparative experimental results between the proposed model and state-of-the-art methods. It can be observed that our proposed FAHCD-Net can achieve an $\text{NME}_\mathrm{io}$ of 2.51 on the 300W Common Subset,  outperforming other methods \cite{Huang2021ADNetLE, Zhu2022OcclusionrobustFA,Xia2022SparseLP,Zhou2023STARLR,Liang2024GeneralizableFL,Gao2024SelfSupervisedFR}. On the 300W Fullset, it can attaine an $\text{NME}_\mathrm{io}$ of 2.99, demonstrating commendable performance. Furthermore, on the AFLW-frontal subset, FAHCD-Net achieved an $\text{NME}_\mathrm{diag}$ of 1.17 (Tab.\ref{tab:aflw_sota}), outperforming other methods \cite{Wan2024PreciseFL,Kumar2020LUVLiFA,Wu2018LookAB,Wang2019DeepHR,Dong2018StyleAN,Yang2017StackedHN}. These results demonstrate that the FAHCD-Net model achieves a relatively advanced level of performance under normal circumstances. This can be attributed to the characteristics of the diffusion model underlying FAHCD-Net. Under normal circumstances, the diffusion model fully leverages its ability to learn data distributions, progressively generating high-quality heatmaps and effectively FLD.

\begin{table}[t]
\centering
\caption{\textbf{Comparisons with state-of-the-art methods on the 300W dataset.}
The error (NME) is normalized by the inter-ocular distance.
$\circ$ and $\diamond$ denote heatmap regression and coordinate regression methods, respectively.}
\renewcommand\arraystretch{1}
\setlength{\tabcolsep}{6pt}

\begin{tabular}{l|ccc}
\hline
\textbf{Method} & \textbf{Common} & \textbf{Challenging} & \textbf{Full} \\
\hline

$\circ$ LAB (CVPR18) \cite{Wu2018LookAB}& 2.98 & 5.19 & 3.49 \\
$\circ$ AWing (ICCV19) \cite{Feng2017WingLF}& 2.72 & 4.52 & 3.07 \\
$\circ$ LUVI (CVPR20) \cite{Kumar2020LUVLiFA}& 2.76 & 5.16 & 3.23 \\
$\circ$ ADNet (ICCV21)\cite{Huang2021ADNetLE} & 2.53 & 4.58 & 2.93 \\
$\circ$ STAR (CVPR23)\cite{Zhou2023STARLR} & 2.52 & 4.32 & 2.87 \\

\hline

$\diamond$ ODN (CVPR19)\cite{odn} & 3.56 & 6.67 & 4.17 \\
$\diamond$ DAG (ECCV20)\cite{dag} & 2.62 & 4.77 & 3.04 \\
$\diamond$ LGSA (TMM21)\cite{tmm9082841} & 2.92 & 5.16 & 3.36 \\
$\diamond$ PIPNet (IJCV21) \cite{pipnetJLS21}& 2.78 & 4.89 & 3.19 \\
$\diamond$ SLPT (CVPR22)\cite{Xia2022SparseLP} & 2.75 & 4.90 & 3.17 \\
$\diamond$ GlomFace (CVPR22)\cite{glomfacezhu2022occlusion} & 2.79 & 4.87 & 3.20 \\
$\diamond$ DTLD (CVPR23) \cite{DTLD}& 2.59 & 4.50 & 2.96 \\
$\diamond$ ATF (TMM23) \cite{9749863_atf_tmm}& 2.75 & 4.89 & 3.17 \\
$\diamond$ EfficientFAN (TNNLS23)\cite{efficientfan} & 2.98 & 5.21 & 3.42 \\
$\diamond$ PicasoNet (TNNLS23)\cite{PicassoNet} & 3.03 & 5.81 & 3.58 \\
$\diamond$ Lite-HRNet (ICIP23)\cite{Lite-HRNet} & 3.97 & 6.89 & 4.54 \\
$\diamond$ Liang et al. (CVPR24)\cite{Liang2024GeneralizableFL} & 2.68 & 4.86 & 3.10 \\

\hline

$\diamond$ \textbf{FAHCD-Net (ours)} & \textbf{2.51} & \textbf{4.49} & \textbf{2.99} \\

\hline
\end{tabular}
\label{tab:300w_sota}
\vspace{-0.5em}
\end{table}

\begin{table}[t]
\centering
\caption{\textbf{Comparisons with state-of-the-art methods on the COFW dataset.}
The error (NME) is normalized by the inter-pupil distance.
$\circ$ and $\diamond$ denote heatmap regression and coordinate regression methods, respectively.}
\renewcommand\arraystretch{1}
\setlength{\tabcolsep}{16pt}

\begin{tabular}{l|cc}
\hline
\textbf{Method} & $\textbf{NME}_\mathrm{ip}$ & $\textbf{FR}_\mathrm{ip}^{10}$ \\
\hline

$\circ$ LAB (CVPR18)\cite{Wu2018LookAB} & 5.58 & 2.76 \\
$\circ$ DCFE (ECCV18)\cite{Valle2018ADC} & 5.27 & 7.29 \\
$\circ$ AWing (ICCV19)\cite{Feng2017WingLF} & 4.94 & 0.99 \\
$\circ$ ADNet (ICCV21)\cite{Huang2021ADNetLE} & 4.68 & 0.59 \\
$\circ$ STAR (CVPR23)\cite{Zhou2023STARLR} & 4.62 & 0.79 \\
$\circ$ DSAT (PR24)\cite{Wan2024PreciseFL} & 4.74 & 0.00 \\

\hline

$\diamond$ ODN (CVPR19)\cite{odn} & 5.30 & 1.78 \\
$\diamond$ MMDN (TNNLS22)\cite{mmdn} & 5.01 & 1.78 \\
$\diamond$ GlomFace (CVPR22)\cite{glomfacezhu2022occlusion} & 4.37 & 1.56 \\
$\diamond$ SLPT (CVPR22) \cite{Xia2022SparseLP}& 4.79 & 1.18 \\
$\diamond$ DSLPT-R50 (TPAMI23)\cite{Ramesh2022HierarchicalTI} & 4.81 & 1.18 \\

\hline

\textbf{$\diamond$FAHCD-Net (ours)} & \textbf{4.65} & \textbf{0.20} \\

\hline
\end{tabular}
\label{tab:cofw_sota}
\vspace{-0.5em}
\end{table}

\begin{table*}[t]
  \centering
  \scriptsize
  \caption{\textbf{Comparisons with state-of-the-art methods on the WFLW dataset.}
  NME is normalized by the inter-ocular distance.
  $\circ$ and $\diamond$ denote heatmap regression and coordinate regression methods, respectively.}
  \vspace{-1em}
  \renewcommand\arraystretch{1}
  \resizebox{\textwidth}{!}{%
  \begin{tabular}{p{3.2cm}|ccccccc}
      \toprule[1pt]
      Method 
      & Testset & Pose Subset & Expression Subset & Illumination Subset & Make-Up Subset & Occlusion Subset & Blur Subset \\
      \midrule

      \makebox[2.8cm][l]{$\circ$\, LAB (CVPR18)\cite{Wu2018LookAB}} & 5.27 & 10.24 & 5.51 & 5.23 & 5.15 & 6.79 & 6.32 \\
      \makebox[2.8cm][l]{$\circ$\, Wing (CVPR18)\cite{Feng2017WingLF}} & 4.99 & 8.43 & 5.21 & 4.88 & 5.26 & 6.21 & 5.81 \\
      \makebox[2.8cm][l]{$\circ$\, HRNet (TPAMI20)\cite{Wang2019DeepHR}} & 4.60 & 7.86 & 4.78 & 4.57 & 4.26 & 5.42 & 5.36 \\
      \makebox[2.8cm][l]{$\circ$\, AWing (ICCV19)\cite{Wang2019AdaptiveWL}} & 4.36 & 7.38 & 4.58 & 4.32 & 4.27 & 5.19 & 4.96 \\
      \makebox[2.8cm][l]{$\circ$\, LUVLi (CVPR20)\cite{Kumar2020LUVLiFA}} & 4.37 & 7.56 & 4.77 & 4.30 & 4.33 & 5.29 & 4.94 \\
      \makebox[2.8cm][l]{$\circ$\, ADNet (ICCV21)\cite{Huang2021ADNetLE}} & 4.14 & 6.96 & 4.38 & 4.09 & 4.05 & 5.06 & 4.79 \\
      \makebox[2.8cm][l]{$\circ$\, SLPT (CVPR22)\cite{Xia2022SparseLP}} & 4.12 & 6.99 & 4.37 & 4.02 & 4.03 & 5.01 & 4.79 \\
      \makebox[2.8cm][l]{$\circ$\, LDEQ (CVPR23)\cite{Micaelli2023RecurrenceWR}} & 3.92 & 6.86 & 3.94 & 4.17 & 3.75 & 4.77 & 4.59 \\
      \makebox[2.8cm][l]{$\circ$\, FRA (CVPR24)\cite{Gao2024SelfSupervisedFR}} & 4.11 & - & - & - & - & - & - \\

      \midrule

      \makebox[2.8cm][l]{$\diamond$\, DAG (ECCV20)\cite{dag}} & 4.21 & 7.36 & 4.49 & 4.12 & 4.05 & 4.98 & 4.82 \\
      \makebox[2.8cm][l]{$\diamond$\, PIPNet (IJCV21)\cite{pipnetJLS21}} & 4.31 & - & - & - & - & - & - \\
      \makebox[2.8cm][l]{$\diamond$\, MMDN (TNNLS22)\cite{mmdn}} & 4.87 & 7.71 & 4.79 & 4.61 & 4.72 & 6.17 & 5.72 \\
      \makebox[2.8cm][l]{$\diamond$\, GlomFace (CVPR22)\cite{glomfacezhu2022occlusion}} & 4.81 & 8.71 & - & - & - & 5.14 & - \\
      \makebox[2.8cm][l]{$\diamond$\, DTLD (CVPR22)\cite{DTLD}} & 4.08 & - & - & - & - & - & - \\
      \makebox[2.8cm][l]{$\diamond$\, ATF (TMM23)\cite{9749863_atf_tmm}} & 4.50 & 7.54 & 4.65 & 4.45 & 4.20 & 5.30 & 5.19 \\
      \makebox[2.8cm][l]{$\diamond$\, EfficientFAN (TNNLS23)\cite{efficientfan}} & 4.54 & 8.20 & 4.87 & 4.39 & 4.54 & 5.42 & 5.04 \\
      \makebox[2.8cm][l]{$\diamond$\, PicasoNet (TNNLS23)\cite{PicassoNet}} & 4.82 & 8.61 & 5.14 & 4.73 & 4.68 & 5.91 & 5.56 \\
      \makebox[2.8cm][l]{$\diamond$\, Lite-HRNet (ICIP23)\cite{Lite-HRNet}} & 5.58 & 9.79 & 6.13 & 5.44 & 5.87 & 6.57 & 6.05 \\

      \hline
      \makebox[2.8cm][l]{$\diamond$\, \textbf{FAHCD-Net (ours)}} &
      \textbf{4.02} & \textbf{7.09} & \textbf{4.59} & \textbf{4.25} & \textbf{4.10} & \textbf{5.18} & \textbf{4.70} \\
      \bottomrule[1pt]
  \end{tabular}%
  }
  \label{tab:wflw_sota}
  \vspace{-2em}
\end{table*}

\begin{table}[t]
  \centering
  \scriptsize
  \caption{\textbf{Comparisons with state-of-the-art methods on the AFLW dataset.}
  $\circ$ and $\diamond$ denote heatmap regression and coordinate regression methods, respectively.}
  \renewcommand\arraystretch{1}
  \resizebox{\columnwidth}{!}{%
  \begin{tabular}{p{2.6cm}|ccc}
      \toprule[1pt]
      Method & $\textbf{NME}_{\mathrm{diag}}$ (full) & $\textbf{NME}_{\mathrm{diag}}$ (frontal) & $\textbf{NME}_\mathrm{box}$ (full) \\
      \midrule

      \makebox[2.6cm][l]{$\circ$\, SHN (CVPR17)\cite{Yang2017StackedHN}} & 2.46 & 1.92 & 3.67 \\
      \makebox[2.6cm][l]{$\circ$\, SAN (CVPR18)\cite{Dong2018StyleAN}} & 1.91 & 1.85 & 4.04 \\
      \makebox[2.6cm][l]{$\circ$\, LAB (CVPR18)\cite{Wu2018LookAB}} & 1.85 & 1.62 & - \\
      \makebox[2.6cm][l]{$\circ$\, HRNet (TPAMI20)\cite{Wang2019DeepHR}} & 1.57 & 2.46 & - \\
      \makebox[2.6cm][l]{$\circ$\, Wing (CVPR17)\cite{Feng2017WingLF}} & - & - & 3.56 \\
      \makebox[2.6cm][l]{$\circ$\, LUVLi (CVPR20)\cite{Kumar2020LUVLiFA}} & 1.39 & 1.19 & 2.28 \\
      \makebox[2.6cm][l]{$\circ$\, DSAT (PR24)\cite{Wan2024PreciseFL}} & 1.35 & 1.17 & - \\

      \midrule

      \makebox[2.6cm][l]{$\diamond$\, DTLD+ (CVPR22)\cite{Li2022TowardsAF}} & 1.37 & - & - \\
      \makebox[2.6cm][l]{$\diamond$\, DSLPT (TPAMI23)\cite{Ramesh2022HierarchicalTI}} & 1.38 & - & - \\

      \hline
      \makebox[2.6cm][l]{$\diamond$\, \textbf{FAHCD-Net (ours)}} &
      \textbf{1.30} & \textbf{1.17} & \textbf{2.08} \\
      \bottomrule[1pt]
  \end{tabular}%
  }
  \label{tab:aflw_sota}
  \vspace{-2em}
\end{table}

\subsection{Evaluation of Robustness against Occlusion}
Under normal conditions, most models perform well. However, when facial images are heavily occluded, many existing methods struggle to maintain their effectiveness. To assess the performance of our FAHCD-Net under these conditions, we conducted experiments on datasets that contain significant occlusions, such as the COFW dataset, the 300W Challenging Subset, and the WFLW dataset.
 
On COFW dataset,  FAHCD-Net achieves an $\text{NME}_\mathrm{ip}$ of 4.65 (Tab. \ref{tab:cofw_sota}), outperforming most other methods \cite{Wan2024PreciseFL,Zhou2023STARLR,Xia2022SparseLP,Huang2021ADNetLE,Wang2019AdaptiveWL,Valle2018ADC,Wu2018LookAB,wan16}. On the 300W Challenging Subset, FAHCD-Net attained an $\text{NME}_\mathrm{io}$ of 4.49, as shown in Tab. \ref{tab:300w_sota}, also surpassing state-of-the-art methods \cite{Gao2024SelfSupervisedFR,Liang2024GeneralizableFL,Zhou2023STARLR,Xia2022SparseLP,Zhu2022OcclusionrobustFA,Huang2021ADNetLE,Wang2019AdaptiveWL,Kumar2020LUVLiFA}. Furthermore, as presented in Tab. \ref{tab:wflw_sota}, FAHCD-Net exhibited exceptional performance on the Make-Up Subset and Occlusion Subset of the WFLW dataset.
These outstanding performances can be attributed to FAHCD-Net's strong robustness to noise and its exceptional handling of frequency information. Specifically, FAHCD-Net enhances noise resistance through the generative process of the diffusion model while effectively extracting structural information from low-frequency signals using the HFA module. These attributes collectively enable FAHCD-Net to excel in occlusion scenarios. 

\subsection{Evaluation of Robustness against Large Poses and Expressions}

Facial images with large poses and expressions present significant challenges in facial landmark detection tasks. To evaluate the proposed model's detection capability under these circumstances, experiments are conducted on the AFLW-full dataset, WFLW dataset and 300W Challenging Subset.

On the AFLW-full dataset, our proposed FAHCD-Net can achieve an $\text{NME}_\mathrm{diag}$ of 1.30 and an $\text{NME}_\mathrm{box}$ of 2.08, as shown in Tab. \ref{tab:aflw_sota}, outperforming other methods \cite{Wan2024PreciseFL,Li2022TowardsAF,Kumar2020LUVLiFA,Feng2017WingLF,wan17,wan18}. On the 300W Challenging Subset, as shown in Tab. \ref{tab:300w_sota}, FAHCD-Net can attain an $\text{NME}_\mathrm{io}$ of 4.49, demonstrating great performance. Additionally, on the WFLW dataset's Pose Subset and Expression Subset, as presented in Tab. \ref{tab:wflw_sota}, FAHCD-Net can also exhibite exceptional results. These results indicate that FAHCD-Net, leveraging the capabilities of the diffusion model to learn statistical properties and simulate the initial data distribution, effectively adapts to variations in facial poses. Meanwhile, the HFA module further utilizes frequency information to capture global structural characteristics from low-frequency information. This enables FAHCD-Net to maintain accurate FLD in scenarios involving pose and expression variations.

\subsection{Evaluation of Robustness against Illumination and Blur}
This part focuses on facial images with varying illumination and blur. To evaluate the proposed model's detection capabilities under these conditions, experiments are conducted on WFLW dataset and the 300W Challenging Subset. On the 300W Challenging Subset, as shown in Tab. \ref{tab:300w_sota}, FAHCD-Net can achieve an $\text{NME}_\mathrm{io}$ of 4.29, demonstrating leading performance. Additionally, on the WFLW dataset's Illumination Subset and Blur Subset, as presented in Tab. \ref{tab:wflw_sota}, FAHCD-Net also exhibites outstanding results. This improvement performance can be attributed to the HFA module and SR loss effectively suppressing high-frequency noise present in the data and generated by the diffusion model. By aligning the frequency distribution of the generated heatmaps with the ground-truth, the module ensures the generation of clean and accurate landmark heatmaps.

\begin{table}[t]
\centering
\scriptsize
\caption{\textbf{Comparison of different conditional heatmaps on the 300W challenging subset.}
NME is normalized by the inter-ocular distance.}
\renewcommand\arraystretch{1}
\resizebox{\columnwidth}{!}{
\begin{tabular}{p{2.4cm}|ccc}
\hline
\textbf{Method} & Stage I & Stage II & Stage III \\
\hline
Mean Shape & 4.81 & 4.73 & 4.69 \\
\textbf{FAHCD-Net (ours)} & \textbf{4.48} & \textbf{4.33} & \textbf{4.29} \\
\hline
\end{tabular}
}
\label{tab5}
\vspace{-1em}
\end{table}

\subsection{Self Evaluation}

\subsubsection{Mean shape as Conditional Heatmap}
To verify whether the proposed FAHCD-Net relies heavily on the quality of the conditional heatmap, we replace the SHN-generated heatmaps~\cite{Yang2017StackedHN} with a mean shape prior, which is obtained by averaging a large number of facial landmark annotations. 

Tab.~\ref{tab5} reports the results on the 300W Challenging subset. Replacing the learned heatmaps with the mean shape leads to a noticeable performance drop. However, the degradation remains moderate. This indicates that the proposed framework does not strictly rely on high-quality heatmap initialization. These results can be attributed to the fact that, although the mean shape lacks instance-specific structural details, it still provides a coarse yet stable spatial prior, which can guide the model toward reasonable landmark localization. Meanwhile, this alternative significantly simplifies the training pipeline and removes the reliance on an additional heatmap generator, making it more practical in real-world scenarios where such prior models are unavailable.




\begin{figure}[t]
	\centering
	\includegraphics[width=\columnwidth]{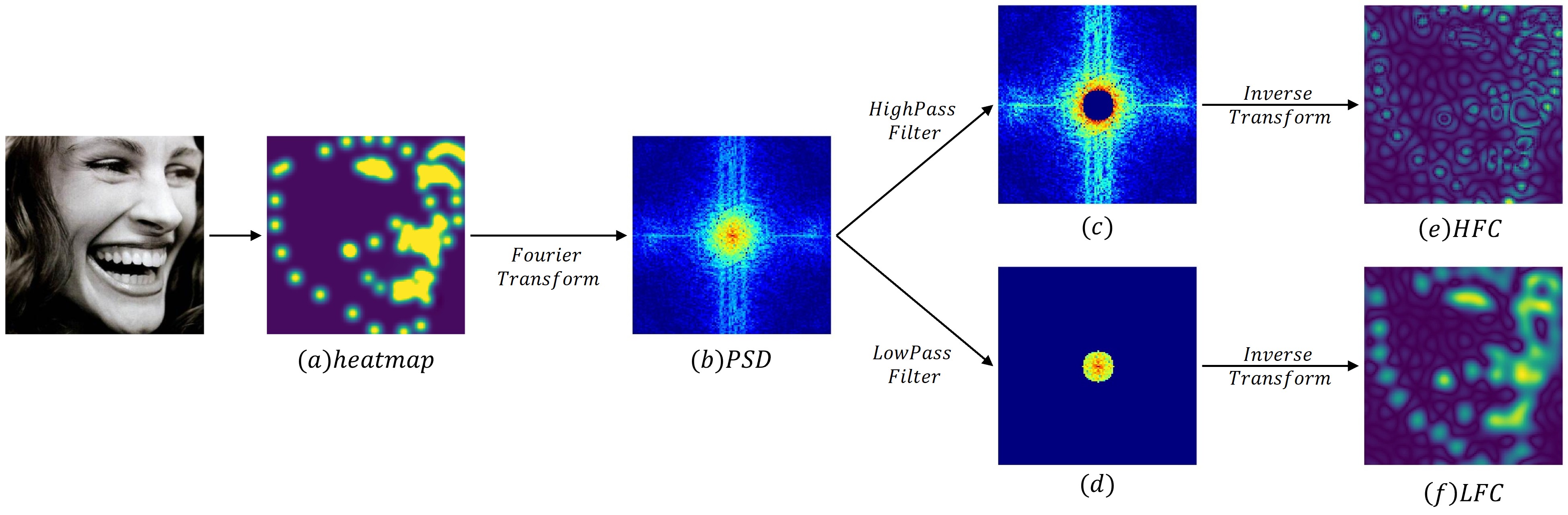}
	\centering
    	\vspace{-2em}
	\caption{The process of frequency domain visualization and reconstruction using the Fourier Transform.
	}
	\label{fig6}
	\vspace{-1em}
\end{figure}

\subsubsection{Frequency Component Analysis}
In this section, we evaluate the role of frequency decomposition and reconstruction in improving the model's detection accuracy. We aim to validate the impact of these operations from the following perspectives.

\textbf{Frequency Domain Visualization.} To better understand the influence of frequency components on model performance, we first apply a \textbf{Fourier Transform} to both the input and output heatmaps, converting them from the spatial domain to the frequency domain. The Fourier Transform reveals the distribution of the image across different frequencies, helping us distinguish between low-frequency smooth areas and high-frequency detailed regions in the image. The Fourier Transform formula is given by:
\begin{small}
	\begin{equation}
		F_h(u,v)=\iint{h(x,y) e^{-i2\pi(ux+vy)}dxdy}	
	\end{equation}
\end{small}where $h(x,y)$ is the heatmap in the spatial domain, $F_h(u,v)$ is the heatmap in the frequency domain, and $u$ and $v$ are the coordinates in the frequency domain. We then calculate the \textbf{Power Spectral Density (PSD)} to gain further insight into the distribution of frequency components. For better visualization, we apply a logarithmic transformation $\log(1+|F_h(u,v)|^2)$ to the PSD values.

Fig.\ref{fig6}(b) presents the frequency domain visualization results after applying the Fourier Transform. Generally, the central part of the image corresponds to low-frequency components, reflecting the smooth regions of the image (such as facial contours, skin, and large structures). The edges of the image correspond to high-frequency components, highlighting fine details (e.g., the contours of the eyes, mouth, and subtle facial textures).

To investigate the influence of different frequency components on image details, we employ low-pass and high-pass filters on the frequency-domain representations of images obtained through Fourier transformation. The effects are then analyzed by applying an inverse Fourier Transform to reconstruct the filtered images.

As shown in Fig.\ref{fig6}(e) and (f), the low-pass filter removes high-frequency information, such as fine details and edges, retaining primarily smooth, low-frequency components. This results in images with diminished sharpness and a lack of intricate details. Conversely, the high-pass filter suppresses low-frequency components while preserving high-frequency details, such as textures and edges, thereby enhancing the prominence of fine structures and sharp transitions in the image. This comparative analysis demonstrates the distinct roles of low- and high-frequency components in defining image features.

These operations help us analyze the influence of frequency components on the heatmap generation process. They also validate the critical role that frequency domain information plays in enhancing the accuracy of landmark detection. By identifying the effects of different frequency bands on the heatmap, we gain insights into how specific frequency components contribute to model performance in facial landmark detection.

\textbf{Frequency Correlation Analysis.} To further evaluate the relationship between frequency components and facial landmark localization accuracy, we quantify the energy of the frequency components and analyze its correlation with the model's performance. We use \textbf{frequency energy} as a metric to quantify the frequency components, which is calculated as follows:

\begin{small}
	\begin{equation}
		E=\sum_{u,v}|F_h(u,v)|^2
	\end{equation}
\end{small}where $E$ represents the energy of the frequency components.

In the experiment, to distinguish between high-frequency and low-frequency energy, we categorize the frequency components based on their position in the frequency spectrum. Specifically, we calculate the distance $D(u,v)$ of the frequency component $(u,v)$ from the center of the spectrum:

\begin{small}
	\begin{equation}
		D(u,v)=\sqrt{u^2+v^2}
	\end{equation}
\end{small}

Next, using a predefined distance threshold $D_{\text{threshold}}$, we classify the frequency components into high-frequency and low-frequency components. In this experiment, the threshold is set to $20\%$ of the spectrum size. Based on this criterion, we divide the spectrum into high-frequency and low-frequency regions, and compute the corresponding energies for each:

\begin{small}
	\begin{align}
		E_\text{low}=\sum_{D(u,v)\le D_\text{threshold}}|F_h(u,v)|^2 \\
		E_\text{high}=\sum_{D(u,v)> D_\text{threshold}}|F_h(u,v)|^2 
	\end{align}
\end{small}

\begin{figure}[!t]
	\centering
	\includegraphics[width=\columnwidth]{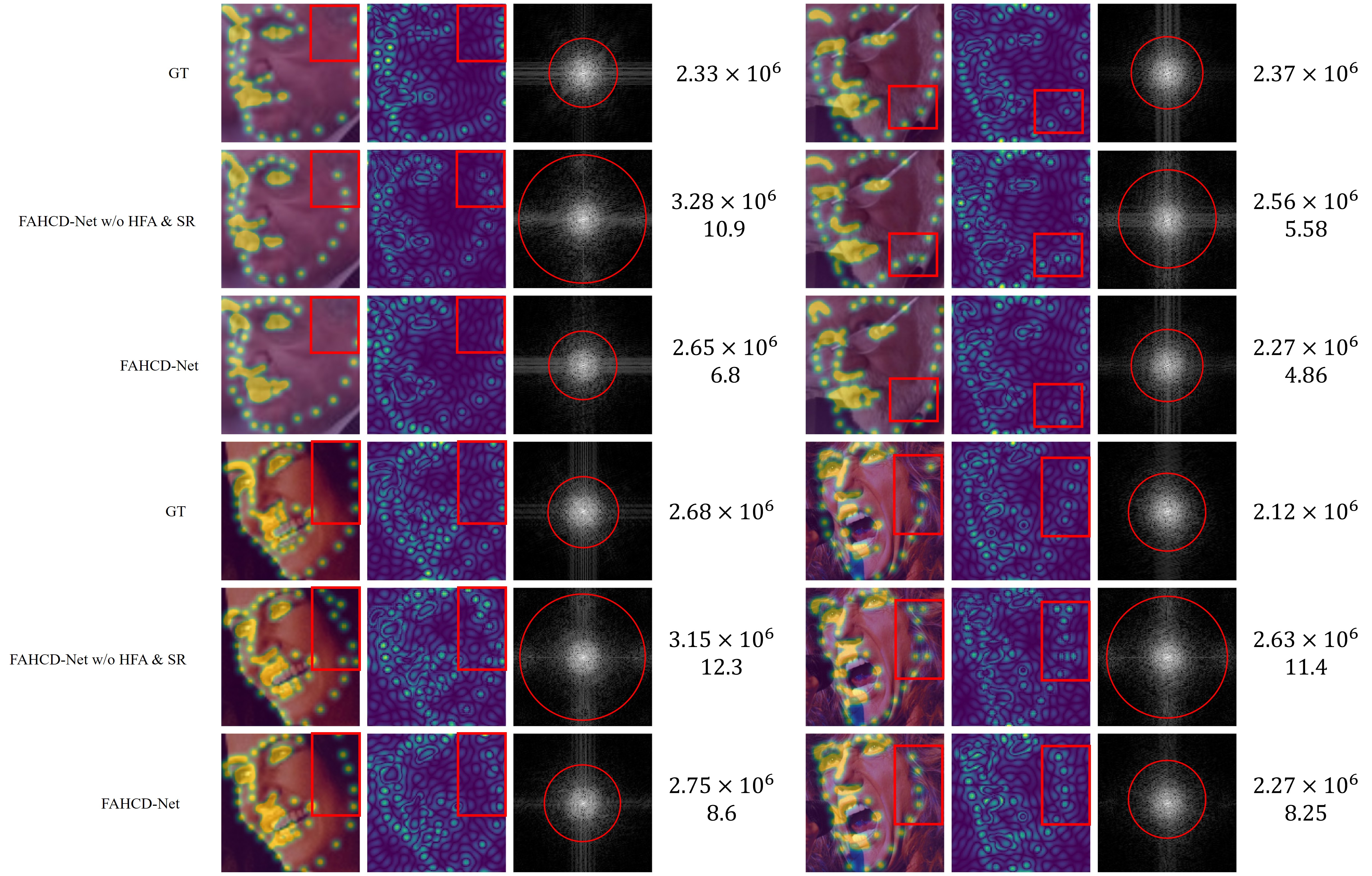}
	\centering
    \vspace{-1.5em}
	\caption{ 
		Results of frequency component analysis. 
		In each group of images, the first, second and third columns display the landmark heatmap, high-frequency component map and PSD (Power Spectral Density) map, respectively. And the fourth column contains the numerical values, which represent the energy of the high-frequency components and their corresponding $\text{NME}_\mathrm{io}$ results. From the analysis, it is evident that, in the absence of the HFA module and SR loss, the model generates excessive high-frequency information, leading to inaccuracies in landmark detection. In contrast, FAHCD-Net effectively suppresses redundant high-frequency noise, resulting in more accurate FLD.
	}
	\label{fig7}
	\vspace{-1em}
\end{figure}

After calculating the low-frequency energy $E_{\text{low}}$ and high-frequency energy $E_{\text{high}}$, we compare these values with the localization error in the facial landmark detection task. The experimental results of frequency decomposition, visualization, and energy calculation are shown in Fig.\ref{fig7}. From the figure, it can be observed that when using the diffusion model-based neural network alone (FAHCD model w/o HFA module), the generated images exhibit noticeable high-frequency noise, which interferes with the accurate localization of landmarks, leading to a decrease in accuracy. However, after introducing the HFA module and the SR loss function, the high-frequency noise is effectively suppressed. 

\begin{table}[t]
\centering
\scriptsize
\caption{\textbf{Effect of multi-dataset training on the 300W challenging subset.}
NME is normalized by the inter-ocular distance.}
\renewcommand\arraystretch{1}
\resizebox{\columnwidth}{!}{
\begin{tabular}{c|ccc|c}
\hline
\textbf{Idx} & 300W & +COFW & +WFLW+AFLW & $\textbf{NME}_{\mathrm{io}}$ \\
\hline
(1) & \checkmark &  &  & 4.76 \\
(2) & \checkmark & \checkmark &  & 4.62 \\
(3) & \checkmark & \checkmark & \checkmark & \textbf{4.49} \\
\hline
\end{tabular}
}
\label{tab7}
\vspace{-1em}
\end{table}

\begin{table}[t]
\centering
\scriptsize
\caption{\textbf{Effect of different regularization terms on the 300W dataset.}
NME is normalized by the inter-ocular distance.}
\renewcommand\arraystretch{1}
\resizebox{\columnwidth}{!}{
\begin{tabular}{p{2.2cm}|ccc}
\hline
Reg. Item & Common & Challenging & Full \\
\hline
HGR & 2.59 & 4.35 & 2.93 \\
TVR & \textbf{2.50} & \textbf{4.29} & \textbf{2.85} \\
\hline
\end{tabular}
}
\label{loss}
\vspace{-1em}
\end{table}


The HFA module adaptively adjusts the balance of different frequency levels, significantly reducing unnecessary high-frequency components, while the SR loss further ensures that the generated landmark heatmaps are smoother and more continuous. After these optimizations, the model successfully improves landmark localization accuracy while managing high-frequency noise, indicating that controlling frequency components plays a crucial role in enhancing model performance. 

\subsubsection{Different Smooth Regularization Item}
To validate the effectiveness of different smooth regularization items, we employ TV regularization and HG regularization as $\mathcal{L}_{sr}$ respectively on 300W dataset. The experimental result are shown in Tab. \ref{loss}. 

Experimental results demonstrate that TV regularization significantly outperforms HG regularization. This is likely because, when generating facial landmark heatmaps using diffusion models, TV regularization more effectively balances noise suppression and detail preservation, maintaining sharp edges and local peaks in key areas. In contrast, HG regularization, due to the incorporation of higher-order gradient information, tends to over-smooth, thereby blurring crucial details.

\subsection{Ablation Study}

In this section, we perform ablation experiment to analyze the contribution of each component to the performance of the FAHCD-Net. The experiment is conducted on the 300W dataset, and evaluation is performed using the Inter-Occular setting.

The experimental results, shown in Tab. \ref{tab7}, demonstrate that the introduction of the HFA module and SR loss significantly enhances the model's robustness and prediction accuracy. By comparing the experimental outcomes under different configurations, we are able to clearly identify the impact of each design component on the performance of proposed FAHCD-Net model.

\section{Conclusion}
Robust and accurate facial landmark detection (FLD) in complex scenarios remains a significant challenge due to structural variations, information loss, and noise interference. In this paper, we propose an innovative FAHCD-Net to address these issues by integrating the FAHCD model and SR loss in a cascaded manner. In each FAHCD model, the HFA module dynamically adjusts the frequency components to better align with the ground-truth, effectively suppressing over-recovered details and avoiding unnecessary high-frequency noise. Additionally, the SR loss is designed to mitigate redundant high-frequency noise while enforcing smoothness and continuity in the generated heatmaps, thereby ensuring accurate FLD. Experimental results on challenging datasets demonstrate that FAHCD-Net exhibits strong robustness and accuracy in complex FLD tasks. Furthermore, our approach shows that adaptive frequency information helps retain facial structure and suppress redundant noise, ultimately improving accuracy. In the future, we aim to develop a universal method for landmark detection across diverse scenarios by combining frequency information with scenario-specific labels.

\ifCLASSOPTIONcaptionsoff
  \newpage
\fi

\bibliographystyle{IEEEtran}
\bibliography{FAHCD-Net}

\end{document}